\documentclass{article}

\usepackage[preprint, nonatbib]{neurips_2024}

\usepackage[utf8]{inputenc} 
\usepackage[T1]{fontenc}    
\usepackage{hyperref}       
\usepackage{amsmath}
\usepackage{arydshln}
\usepackage{cleveref}
\usepackage{wrapfig}
\crefname{section}{Sec.}{Secs.}
\Crefname{section}{Section}{Sections}
\Crefname{table}{Table}{Tables}
\crefname{table}{Tab.}{Tabs.}
\Crefname{table}{Tab.}{Tabs.}
\crefname{figure}{Fig.}{Figs.}
\Crefname{figure}{Fig.}{Figs.}
\usepackage{url}            
\usepackage{booktabs}       
\usepackage{amsfonts}       
\usepackage{nicefrac}       
\usepackage{microtype}      
\usepackage{xcolor}         
\usepackage{natbib}
\usepackage{enumitem}
\setcitestyle{numbers,square}
\usepackage{graphicx}
\usepackage{geometry}
\usepackage{colortbl}
\usepackage{makecell}
\usepackage{dsfont}
\usepackage{changepage}
\usepackage{hhline}
\definecolor{aliceblue}{rgb}{0.94, 0.97, 1.0}
\usepackage{subcaption}
\usepackage{xspace}
\usepackage{stfloats}
\usepackage{pifont}
\usepackage{fontawesome}
\usepackage{multirow}
\usepackage{comment}
\newcommand*{\eg}{\textit{e.g.}\@\xspace}
\newcommand*{\ie}{\textit{i.e.}\@\xspace}

\newcommand*{\etal}{\textit{et al.}\@\xspace}

\definecolor{green_im}{rgb}{0.0, 0.5, 0.0}
\definecolor{red_im}{rgb}{0.5, 0.0, 0.0}

\title{EmoLLM: Multimodal Emotional Understanding Meets Large Language Models}

\author{%
  Qu Yang ~ ~ Mang Ye~\footnotemark[1] ~ ~ ~ Bo Du \\
School of Computer Science, Wuhan University, Wuhan, China.\\
  \texttt{\{yangqu,~yemang,~dubo\}@whu.edu.cn} \\
  \texttt{\href{https://github.com/yan9qu/EmoLLM}{https://github.com/yan9qu/EmoLLM}} \\
}

\begin{document}

\renewcommand{\thefootnote}{\fnsymbol{footnote}}
\footnotetext[1]{Corresponding Author}
\renewcommand*{\thefootnote}{\arabic{footnote}}

\maketitle

\begin{abstract}
Multi-modal large language models (MLLMs) have achieved remarkable performance on objective multimodal perception tasks, but their ability to interpret subjective, emotionally nuanced multimodal content remains largely unexplored. Thus, it impedes their ability to effectively understand and react to the intricate emotions expressed by humans through multimodal media. To bridge this gap, we introduce \textbf{EmoBench}, the first comprehensive benchmark designed specifically to evaluate the emotional capabilities of MLLMs across five popular emotional tasks, using a diverse dataset of \textasciitilde287k images and videos paired with corresponding textual instructions. Meanwhile, we propose \textbf{EmoLLM}, a novel model for multimodal emotional understanding, incorporating with two core techniques. 
1) Multi-perspective Visual Projection, it captures diverse emotional cues from visual data from multiple perspectives. 2) EmoPrompt, it guides MLLMs to reason about emotions in the correct direction. Experimental results demonstrate that EmoLLM significantly elevates multimodal emotional understanding performance, with an average improvement of 12.1\% across multiple foundation models on EmoBench. Our work contributes to the advancement of MLLMs by facilitating a deeper and more nuanced comprehension of intricate human emotions, paving the way for the development of artificial emotional intelligence capabilities with wide-ranging applications in areas such as human-computer interaction, mental health support, and empathetic AI systems. Code, data, and model will be released.
\end{abstract}\vspace{-1em}

\section{Introduction}
\begin{figure}[t]
\centering
\includegraphics[width=0.95\linewidth]{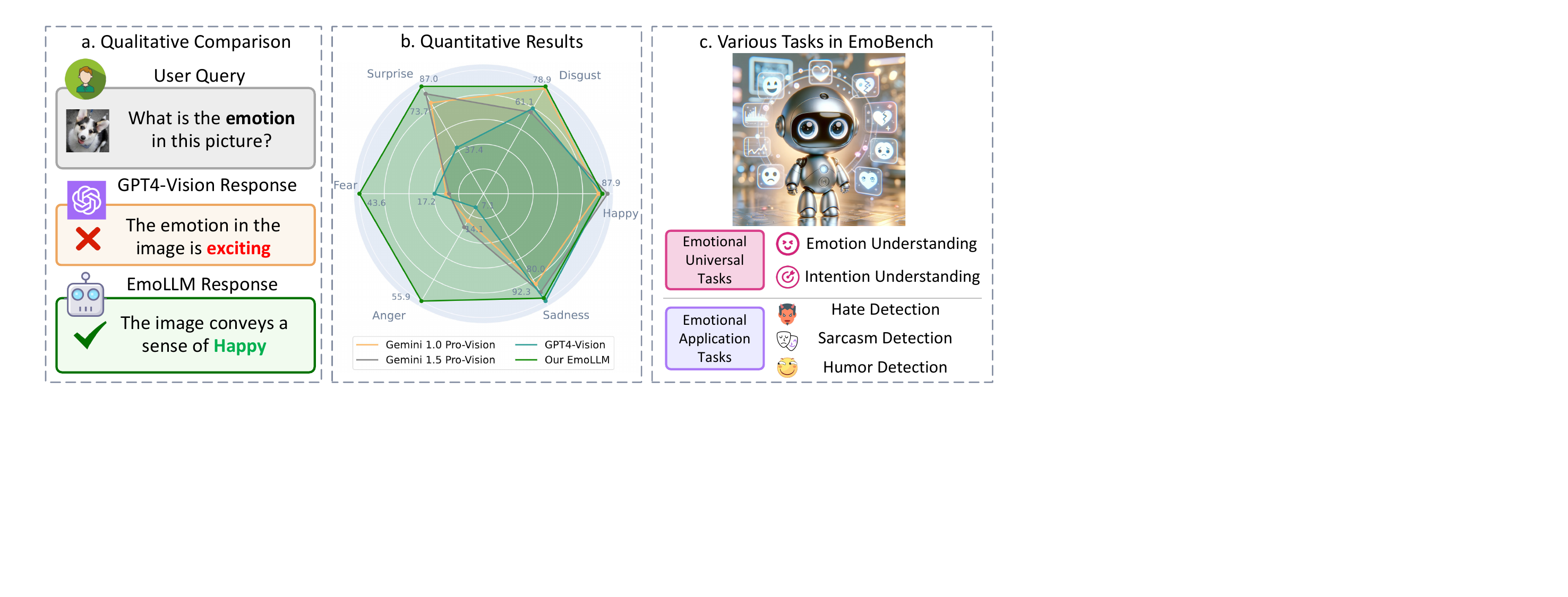}
\vspace{-0.6em}
\caption{Qualitative (a) and quantitative (b) comparison of EmoLLM with GPT4-Vision and other SOTA MLLMs. EmoLLM outperforms other models, particularly in recognizing nuanced emotions such as anger and sadness. (c) Overview of the diverse tasks in EmoBench, including emotional universal tasks, emotional application tasks (hate, sarcasm, and humor detection).}
\label{fig:fig1_intro}\vspace{-1.3em}
\end{figure}

\textit{Do androids dream of electric sheep?} This thought-provoking question from Philip K. Dick's seminal novel underscores a fundamental divide between artificial intelligence and humanity – the capacity for genuine emotion. In our modern era, Multimodal Large Language Models (MLLMs)~\cite{gpt4, llama2, llamaadapter1, gao2023llama-adapter2, dai2024instructblip, mllm1_T5} have achieved remarkable performance, even surpassing human capabilities in domains such as perception and cognition. However, when it comes to the realm of emotions, state-of-the-art MLLMs appear to be lacking in their ability to accurately interpret and respond to emotional cues. While existing MLLMs can generate basic responses to human queries regarding emotional aspects, the accuracy of their responses remains unsatisfactory, especially in nuanced categories such as fear and anger~(\cref{fig:fig1_intro}(b)). Moreover, even LLMs that have been employed for text-based emotional analysis often fall short when confronted with the complexities of multimodal emotional tasks, which require the integration of visual, auditory, and textual cues. A primary factor contributing to this limitation is the scarcity of comprehensive emotional datasets for training MLLMs, as publicly available datasets generally focus on objective visual abilities~\cite{zhang2017s3fd}. This gap not only mirrors the philosophical questions raised by Dick's narrative but also motivates us to explore the vast, uncharted territories of emotional intelligence within MLLMs.

To bridge this gap, we propose \textbf{EmoBench}, a comprehensive benchmark designed to serve two critical functions: providing a rich source of training materials to enhance the performance of MLLMs and evaluating their emotional understanding capabilities. EmoBench encompasses a diverse range of tasks (\cref{fig:fig1_intro}), which we categorize into Universal Emotional Tasks and Emotional Application Tasks. Universal Emotional Tasks include multimodal emotion recognition and intent understanding, both represented as a classification paradigm. Emotional Application Tasks, on the other hand, focus on specific challenges in social media applications, such as Hate, Sarcasm, and Humor Detection. To construct EmoBench, we first collected a diverse dataset for each subtask, as illustrated in \cref{tab:datasets}. Subsequently, we employed GPT-4~\cite{gpt4} to generate a wide array of question templates for each subtask, ultimately compiling a dataset of approximately 287,000 multimodal instructions. By offering a large-scale, diverse, and carefully curated dataset, EmoBench enables rigorous enhancement and evaluation of the emotional understanding capabilities of MLLMs.

With our proposed EmoBench, existing MLLMs can be empowered with better emotional understanding capabilities with downstream fine-tuning. Current MLLMs typically follow a two-step process: modality projection and LLM reasoning. However, these models still struggle to effectively capture and reason about the complex and nuanced emotions present in multimodal data. To address this challenge, we propose \textbf{EmoLLM}, a novel model that incorporates two key techniques: \textbf{Multi-perspective Visual Projection} and \textbf{EmoPrompt}. Multi-perspective Visual Projection captures diverse emotional cues by considering multiple viewpoints. Specifically, we use the features of objects in different feature maps as content information and construct the objects and their relationships as graph-based relational information. By jointly mining these two aspects of information, we can extract features that are more suitable for emotional tasks.

In the reasoning stage, Chain-of-Thought (CoT)~\cite{cot} is a common and effective method. Inspired by CoT, we first let EmoLLM observe objects in multimedia data and then infer emotions based on these observations. However, a significant problem arises, \ie, the correctness of the first-stage observations determines the accuracy of the final inference. To mitigate this issue, EmoPrompt incorporates specific examples stored for the current task. To ensure the correctness of these examples, we present GPT-4V with data samples and ground truth labels to obtain an accurate CoT process. Whenever a prompt is required, EmoPrompt selects one such example to guide the reasoning process.

We summarize our contributions as follows:
\begin{itemize}
\item We introduce EmoBench, a comprehensive benchmark designed to enhance and evaluate the emotional understanding capabilities of MLLMs across a diverse range of tasks, providing a large-scale dataset of~\textasciitilde287k instructions.
\item We propose EmoLLM, which incorporates Multi-perspective Visual Projection to capture diverse emotional cues and EmoPrompt to guide the reasoning process.
\item We conduct extensive experiments on the EmoBench benchmark, demonstrating that EmoLLM achieves substantial improvements over baseline models, with an average improvement of 12.1\% across multiple foundation models.
\end{itemize}

\section{Related Works}
\label{related_works}
\subsection{Multi-modality Emotional Tasks and Methods}
Multimodal emotion recognition, which analyzes feelings through speech, text, and visual cues, has been a growing area of research. Early datasets like IEMOCAP~\cite{iemocap_busso_lre_08} provide vital audiovisual interaction data but are limited by their focus on scripted events and lack of speaker diversity. Subsequent datasets, such as CMU-MOSEI~\cite{MOSEI} and MELD~\cite{poria2019meld}, address these limitations by offering more naturalistic expressions from videos and television shows. Emotic~\cite{kosti2017emotic1,kosti2019emotic2} and GoEmotions~\cite{demszky2020goemotions} further expand the scope of resources for emotion recognition. In the related field of intention understanding, contemporary datasets like CLINC150~\cite{larson2019evaluation}, HWU64~\cite{liu2019benchmarking}, Intentonomy~\cite{jia2021intentonomy}, Snips~\cite{coucke2018snips}, MDID~\cite{kruk-etal-2019-integrating}, MSED~\cite{jia2022beyond}, and BANKING77~\cite{casanueva2020efficient} are derived from a diverse range of sources, including online forums and social media to explore the user intent~\cite{yang2023composed}. The MIntRec dataset~\cite{mintrec_zhanghanlei_mm_22} takes a unique approach by utilizing TV series clips to capture the complex intentions portrayed by actors.

Building upon these datasets, numerous methods~\cite{sqhy1,sqhy2,skl} have been proposed to advance the field of multimodal emotion recognition. Lee~\etal~\cite{tsai2019multimodal} introduce the Multimodal Transformer (MulT), which employs the vanilla Transformer~\cite{Transformer} architecture and directional cross-modal attention to learn effective multimodal language representations. Hazarika~\etal~\cite{hazarika2020misa} propose Modality-Invariant and -Specific Representations (MISA), which differentiates modality features into invariant and specific subspaces to aid in fusion and prediction. Yang~\etal~\cite{MFSA} introduce MFSA, a transformer-based model that leverages adversarial learning to create modality-specific and -agnostic representations for sentiment recognition. Recently, Zhang~\etal~\cite{zhang2023dialoguellm} attempted to use GPT~\cite{gpt4} to convert multimodal emotion tasks into text emotion recognition. However, this approach relies on pre-processing by an MLLM and is not suitable for practical applications.

\subsection{Multi-modality Large Language Models}
Large language models (LLMs), such as GPT-4~\cite{gpt4}, Gemini-Pro~\cite{team2023gemini}, and LLaVA~\cite{liu2024llava}, have demonstrated remarkable language abilities in capturing general knowledge. By incorporating visual and audio inputs into LLMs using techniques like CLIP~\cite{clip} and additional adapting modules~\cite{baevski2020wav2vec, whisper}, multi-modality large language models (MLLMs)~\cite{chatunivi,lyu2023macaw, han2023onellm} have been developed to tackle a variety of multi-modal tasks. These tasks include image captioning~\cite{llmcaption1, llmcaption2}, visual question answering (VQA)~\cite{llmavqa1, llmavqa2}, and other language-related capabilities~\cite{qinstruct}. However, as revealed by our previous research (\cref{fig:fig1_intro} b), the emotional understanding abilities of MLLMs remain unsatisfactory, particularly when dealing with complex emotions such as anger and fear, or emotional categories that require reasoning. We attribute this limitation primarily to the lack of relevant data and specialized models. To address this issue, we introduce \textbf{EmoBench}, the first emotional instruction tuning dataset designed to enhance the emotional understanding capabilities of various MLLMs and enable them to better navigate the realm of emotional comprehension.

\begin{table}[t]
    \small
    \caption{The statistics of various data sources in EmoBench.}
    \label{tab:datasets}
    \centering
    \begin{tabular}{cclcccccc}
    \toprule
    \multirow{2}{*}{\textbf{Category}} & \multirow{2}{*}{\textbf{Sub-task}} &\multirow{2}{*}{\textbf{Dataset}} & \multicolumn{4}{c}{\textbf{Modality}} &  \multicolumn{2}{c}{\textbf{Sampled Size~(k)}}  \\ 
     & &  & \faFileTextO  & \faFileImageO & \faFileMovieO & \faFileAudioO & \textbf{Train} & \textbf{Val \& Test} \\\midrule
\multirow{5}{*}{\shortstack{Universal \\ Emotional\\Tasks}} 
& Emotion & Emotic~\cite{kosti2017emotic1,kosti2019emotic2} &  \ding{55} & \checkmark &  \ding{55}&  \ding{55} &16.2 &6.4 \\
&Emotion & Caer-S~\cite{lee2019caers} &  \ding{55} & \checkmark &  \ding{55}&  \ding{55} & 42.0 & 21.0  \\
&Emotion & Meld~\cite{poria2019meld}& \ding{55} & \ding{55} & \checkmark & \checkmark&11.1 & 2.6 \\
& Emotion & Emotion\_6~\cite{peng2015emotionroi}  &  \ding{55} & \checkmark &  \ding{55}&  \ding{55}  &1.3 & 0.6\\
& Intention & MintRec~\cite{mintrec_zhanghanlei_mm_22} &  \ding{55} & \checkmark &  \ding{55}&  \ding{55} &1.7 & 0.4 \\\midrule
\multirow{3}{*}{\shortstack{Emotional \\ Application\\Tasks}}
& Humor & SMILE~\cite{hyun2023humor_dete} & \ding{55}  & \ding{55} &   \checkmark &   \checkmark &8.6 & 1.0 \\
& Hate & MMHS~\cite{van2023hateful_dete} &  \checkmark & \checkmark &  \ding{55}&  \ding{55} & 139.8 & 10  \\
& Sarcasm & MMSD~\cite{qin2023mmsd2}& \checkmark & \ding{55} & \checkmark & \checkmark & 22.2 &2.4 \\
\bottomrule
    \end{tabular}
\end{table}

\begin{figure}[t]
\centering
\includegraphics[width=\linewidth]{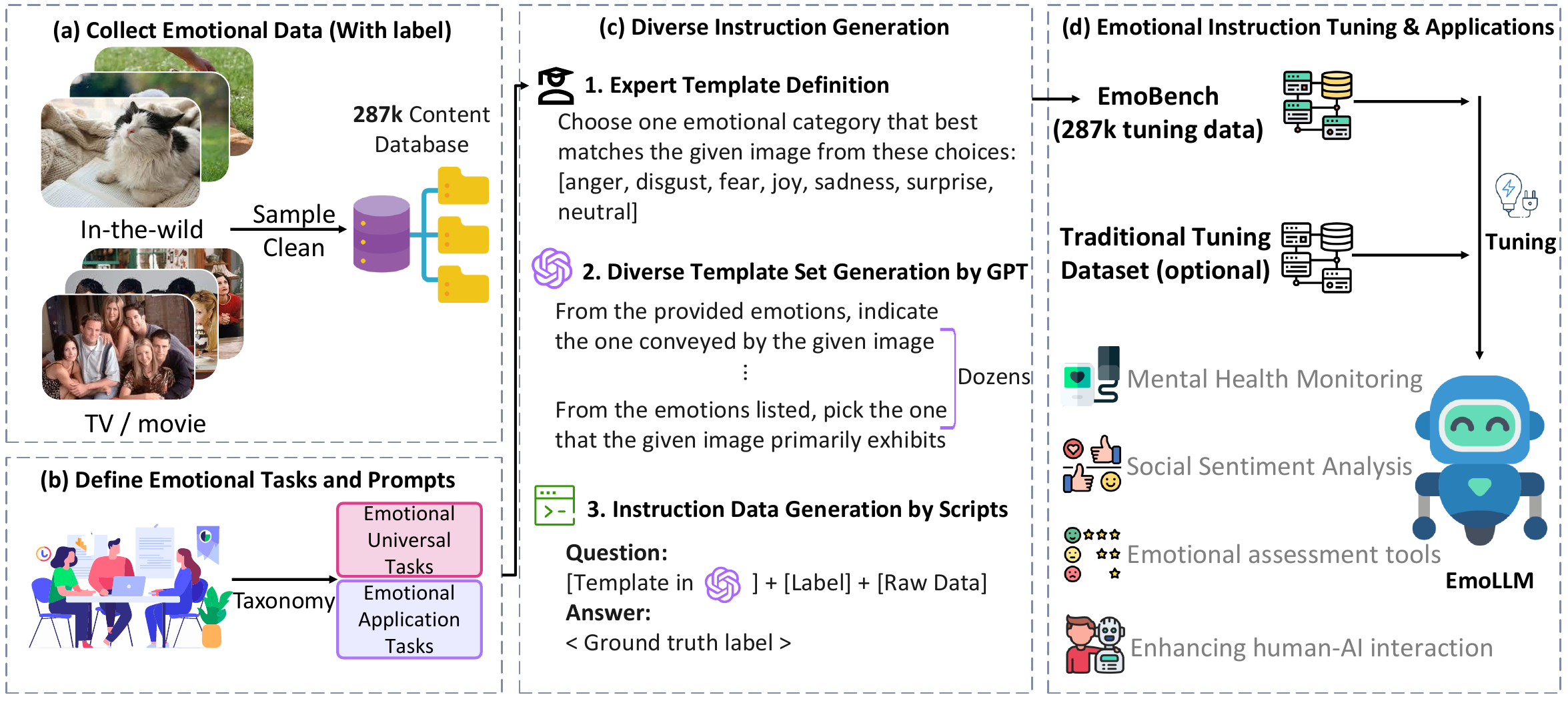}
\caption{Overview of the EmoBench benchmark and its applications. (a) EmoBench is built upon a diverse content database. (c) The process of creating EmoBench involves expert template definition, diverse template set generation, and instruction generation. (d) The proposed EmoLLM is designed to leverage the EmoBench for improving the multi-modal emotional understanding capabilities.}
\label{fig:fig4_instruction}\vspace{-1em}
\end{figure}

\section{EmoBench}
\label{emoBench}
As a cornerstone of emotional tasks, we introduce \textbf{EmoBench}, a pioneering large-scale dataset comprised of conversations focused on emotional dimensions. Initially, we explore the rationale and provide a detailed definition of the tasks associated with \textbf{EmoBench} in~\cref{sec:label_preparation}. Following the task definition, we ensure a diverse and balanced representation of emotional content by sub-sampling from \textbf{eight} distinct datasets. We organize these samples into conversations using generative models and automated scripts, as detailed in~\cref{sec:instruction}.
\subsection{Data Preparation and Task Definition}\label{sec:label_preparation}
The data in \textbf{EmoBench} is sourced from various emotion~\cite{kosti2017emotic1,kosti2019emotic2,poria2019meld,lee2019caers,peng2015emotionroi} and intention~\cite{mintrec_zhanghanlei_mm_22,hyun2023humor_dete,van2023hateful_dete, qin2023mmsd2} datasets. As outlined in~\cref{tab:datasets}, we define two major categories of tasks: universal emotional tasks and emotional application tasks. For the former, which involves multi-modal emotion and intention understanding, we select well-known, data-rich works~\cite{kosti2017emotic1, lee2019caers, poria2019meld, peng2015emotionroi, mintrec_zhanghanlei_mm_22} from the community. From a classification perspective, LLMs are expected to choose the label that best matches the data content. However, considering real-world applications where a predefined label list may not be available, we also explore open-set understanding, where LLMs directly provide the predicted category without a predefined label list.
For emotional application tasks, we identify sub-tasks with significant applications in the industry, particularly those frequently encountered or crucial in social media, such as humor~\cite{hyun2023humor_dete}, hate~\cite{van2023hateful_dete}, and sarcasm~\cite{qin2023mmsd2} detection.
\subsection{Instruction Construction}\label{sec:instruction}
With the assistance of LLMs, data annotation has become increasingly streamlined. We adopt a similar approach and utilize a GPT-participated pipeline to establish a paradigm akin to visual (multimodal) question answering. For the Universal Emotion Tasks outlined in~\cref{sec:label_preparation}, we manually create a question template,~\eg, "Question: Question\_base + [LABEL\_SET]. <DATA> Answer: [LABEL]". The Question\_base is derived from the diverse questions generated by GPT, such as: ``Identify the only emotion depicted in the given image from the following options''; [LABEL\_SET] represents the label set of the current subtask, such as [anger, disgust, fear, joy, sadness, surprise] in the emotion recognition task; <DATA> is the multi-modal data placeholder; and [LABEL] corresponds to the ground-truth label in the original sub-task dataset, reflecting the emotion category of the multi-modal data.
For Emotional Application Subtasks, we modify the question format to a binary choice,~\eg, ``Does the given multi-modal data contain sarcasm? Please answer Yes or No''.

\section{Methodology}
\label{methodology}
In this section, we provide a detailed overview of EmoLLM. We begin by describing the architecture of the model, then delve into each component of EmoLLM.

\begin{figure}[t]
  \centering
  \includegraphics[scale=0.343]{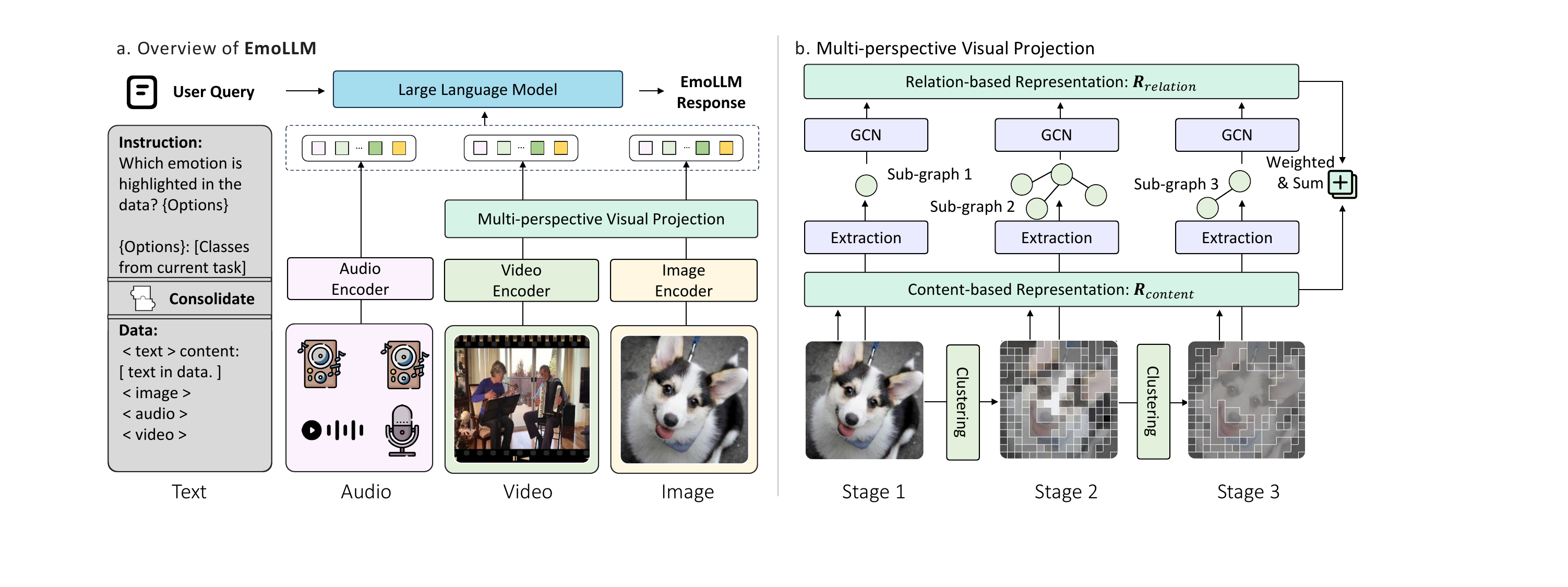}
  \caption{Overview of the EmoLLM framework. (a) EmoLLM takes a user query and multimodal data as input, which are processed by a LLM and modality-specific encoders, respectively. (b) The Multi-perspective Visual Projection consists of various stages, each extracting features from visual tokens and building a graph connecting cluster centers. The combined representations form a comprehensive understanding of the emotional aspects.}
  \label{fig:fig2_overview}
  \vspace{-1em}
\end{figure}

\subsection{Model Overview}
We present an overview of EmoLLM in this section, as shown in~\cref{fig:fig2_overview}. There are three major modules in EmoLLM as follows:

\noindent \textbf{Modality Encoding:} To incorporate additional modalities such as visual and audio data, we integrate extra modality encoders into EmoLLM. This enhancement enables our model to effectively handle multiple modalities.

\noindent \textbf{Multi-perspective Visual Projection:} To effectively capture diverse emotional cues from visual data, we propose the MVP module. Unlike traditional methods that rely on a single perspective, MVP employs a multifaceted approach, analyzing visual data from multiple viewpoints. By constructing a graph-based representation of the relationships between object features, MVP enables EmoLLM to extract a rich set of emotionally relevant features.

\noindent \textbf{EmoPrompt Reasoning:} EmoPrompt leverages the capabilities of GPT-4V~\cite{gpt4} to generate accurate and contextually appropriate prompts. By providing GPT-4V with carefully curated data samples and their corresponding ground truth labels, EmoPrompt facilitates a reliable Chain-of-Thought (CoT) process. This CoT process serves as a blueprint for EmoLLM's reasoning, ensuring that it stays on track and arrives at emotionally coherent conclusions.

\subsection{Modality Encoding}
We design the corresponding modal encoding module for common modalities in emotional tasks, including the following three parts:

\noindent \textbf{Visual Modality Encoding:} To encode visual information, including images and video frames, we employ the CLIP-VIT-L/14 model proposed by Radford~\etal~\cite{clip}. CLIP is a novel framework that learns directly from unprocessed textual data related to images, enabling it to exploit a significantly wider range of supervision. The details of the visual encoding process are described in~\cref{subsec:MVP}.

\noindent \textbf{Audio Modality Encoding:} For encoding audio signals and extracting meaningful representations from audio data, we utilize the WHISPER-BASE model introduced by Radford~\etal~\cite{whisper}. WHISPER is a multilingual speech recognition model trained on a vast audio dataset with weak supervision, making it well-suited for capturing rich information from audio inputs.

\noindent \textbf{Textual Modality Encoding:} Large Language Models (LLMs) are typically pre-trained on massive text corpora, enabling instruction-tuned LLMs to effectively process textual information. In EmoLLM, we use LLaMA2-7B~\cite{llama2} as the foundation model for textual modality encoding, leveraging its strong language understanding capabilities.

Given a video $\boldsymbol{x}_{v} \in \mathbb{R}^{L_{v} \times d_{v}}$, an image $\boldsymbol{x}_{i} \in \mathbb{R}^{L_{i} \times d_{i}}$, an audio signal $\boldsymbol{x}_{a} \in \mathbb{R}^{L_{a} \times d_{a}}$, and a user input text $\boldsymbol{x}_{t} \in \mathbb{R}^{L_{t} \times d_{t}}$, we employ pre-trained models to encode the multimodal features. Specifically, we use the Multi-perspective Visual Projection (MVP) module, to encode the visual features. For the audio signal, we first apply the WHISPER model and then use a multilayer perceptron (MLP) to transform them into the desired dimension. The encoding process can be formulated as follows:
\begin{equation}
        \boldsymbol{h}_{i}=\operatorname{MVP}\left(\boldsymbol{x}_{i}\right), \boldsymbol{h}_{v}=\operatorname{MVP}\left(\boldsymbol{x}_{v}\right), \boldsymbol{h}_{a}=\operatorname{MLP}(\operatorname{WHISPER}\left(\boldsymbol{x}_{a}\right)),
\end{equation}
where $\boldsymbol{h}_{i} \in \mathbb{R}^{L_{i} \times d_{h}}$,$ \boldsymbol{h}_{v} \in \mathbb{R}^{L_{v} \times d_{h}} $~and~$\boldsymbol{h}_{a} \in \mathbb{R}^{L_{a} \times d_{h}}$ denote the encoded image, video, and audio features, respectively. The dimension of the modality-specific features is represented by $d_{h}$

\subsection{Multi-perspective Visual Projection}\label{subsec:MVP}
In this section, we introduce Multi-perspective Visual Projections designed for emotional tasks. We consider two important aspects of multimodal emotional tasks: (1) mining objective object information in multimodal data, which we call \textbf{content-based perspective}, and (2) observing the connections and relationships between objects, which we refer to as \textbf{relation-based perspective}. To better understand the emotional aspects highlighted in the data, we believe that MLLMs should consider both the content-based and relation-based perspectives to deepen their understanding of emotional factors.

Given an input image (or a frame of video) $\boldsymbol{x}_{i}$, we adopt the vision encoder of CLIP~\cite{clip} to extract the original visual tokens $\boldsymbol{Z}=\left\{z_{i}\right\}_{i=1}^{L}$, where $L$ is the number of visual tokens. Following Jin~\etal~\cite{chatunivi}, we then utilize DPC-KNN~\cite{DPC-KNN}, a k-nearest neighbor-based density peaks clustering algorithm, to cluster the visual tokens and obtain the content-based representation. The local density $\rho_i$ and distance index $\delta_i$ of each token $z_{i}$ are computed as follows:
\vspace{-0.5em}
\begin{equation}
\begin{aligned}
\label{eq:2}
\rho_i&=\textrm{exp}\big(-\frac{1}{K}\sum_{z_{k}\in \textrm{KNN}(z_{i}, \boldsymbol{Z})}\Vert z_{k}-z_{i} \Vert^2\big),\
\delta_i&=
\begin{cases}
\min\limits_{j:\rho_j>\rho_i} \Vert z_{j}-z_{i} \Vert^2, & \text{if\ $\exists j$\ s.t.\ $\rho_j>\rho_i$,}\\
\max\limits_{j} \Vert z_{j}-z_{i} \Vert^2, & \text{otherwise,}
\end{cases}
\end{aligned}
\end{equation}
where $\textrm{KNN}(z_{i}, \boldsymbol{Z})$ denotes the K-nearest neighbors of $z_{i}$ in $\boldsymbol{Z}$ after removing $z_{i}$. Tokens with relatively high $\rho_i \times \delta_i$ are identified as cluster centers, and other tokens are allocated to their nearest cluster center based on Euclidean distances. The average token within each cluster represents the corresponding cluster $z^{\prime}_{i}$.

To obtain the relation-based representation, we construct a graph $\mathcal{G}=(\mathcal{V}, \mathcal{E})$ using the cluster centers. Each cluster center $z^{\prime}{i}$ becomes a node $v_i \in \mathcal{V}$, with the feature of each cluster center used as the node's value. To determine the edge weights, we first calculate the Euclidean distance between all cluster centers:
\begin{equation}
d_{ij} = \Vert z^{\prime}_{i} - z^{\prime}_{j} \Vert_2.
\end{equation}
We then normalize the distances to the range [0, 1] using min-max normalization:
\begin{equation}
\tilde{d}_{ij} = \frac{d_{ij} - \min_{i,j}(d_{ij})}{\max_{i,j}(d_{ij}) - \min_{i,j}(d_{ij})}.
\end{equation}
To determine the adjacency matrix $\boldsymbol{A}$, we set a threshold $\tau$ and consider nodes $i$ and $j$ as adjacent if their normalized distance $\tilde{d}_{ij}$ is less than or equal to $\tau$:
\begin{equation}
\boldsymbol{A}{ij} =
\begin{cases}
1, & \text{if } \tilde{d}_{ij} \leq \tau,\\
0, & \text{otherwise.}
\end{cases}
\end{equation}
We apply a multi-layer graph convolutional network (GCN)~\cite{kipf2016semi} to the constructed graph. The graph convolution operation at layer $l$ can be formulated as:
\begin{equation}
\boldsymbol{H}^{(l+1)} = \sigma(\hat{\boldsymbol{D}}^{-\frac{1}{2}}\hat{\boldsymbol{A}}\hat{\boldsymbol{D}}^{-\frac{1}{2}}\boldsymbol{H}^{(l)}\boldsymbol{W}^{(l)}),
\end{equation}
where $\hat{\boldsymbol{A}}=\boldsymbol{A}+\boldsymbol{I}$ is the adjacency matrix with added self-connections, $\hat{\boldsymbol{D}}$ is the degree matrix of $\hat{\boldsymbol{A}}$, $\boldsymbol{H}^{(l)}$ is the feature matrix at layer $l$, $\boldsymbol{W}^{(l)}$ is the trainable weight matrix at layer $l$, and $\sigma$ is the activation function. The output of the last layer $\boldsymbol{H}^{(m)}$ serves as the relation-based representation, where $m$ is the number of layers.

For a video with the $m$-th frame $\boldsymbol{Z}^m=\{z_i^m\}_{i=1}^{L}$, following~\cite{chatunivi}, we apply mean-pooling over all tokens to obtain the frame-level representation $f^m$:
\begin{equation}
f^m = \frac{1}{L}\sum_{i=1}^{L}z_i^m.
\end{equation}
We then use DPC-KNN~\cite{DPC-KNN, chatunivi} to cluster the frames and identify critical events. The set of visual tokens within the $n$-th event $\boldsymbol{F}_{n}$ is denoted as $\tilde{\boldsymbol{Z}}_{n}=\{z_i^m| m \in \boldsymbol{F}_n,\ i \in {1,2,...,L}\}$. To make the visual tokens expand over frames within each event, we adjust the local density and distance index calculations according to~\cref{eq:2}.
The expanded visual tokens are concatenated together in order of events to ensure temporal understanding.
To provide multi-scale visual features, we adopt a three-step aggregation process for each input image or video. The outputs from each merging step are concatenated and transformed using a trainable projection matrix $\boldsymbol{W}$ to obtain the content-based representation $\boldsymbol{R}_{content}$. The relation-based representation $\boldsymbol{R}_{relation}$ is obtained from the aggregation of the GCN output in each stage~$\boldsymbol{H}^{(m)}$. The final feature representation $\boldsymbol{h}_{i}$ is the linear combination of the content-based and relation-based representations with a coefficient~$\alpha$:
\begin{equation}
\boldsymbol{h}_{i} = (\alpha \times \boldsymbol{R}_{content}) \oplus \boldsymbol{R}_{relation}.
\end{equation}
By integrating content-based and relation-based representations, MVP aims to enhance the ability of model to reason about the relationships between visual elements and improve its performance on downstream emotional tasks. The resulting feature representation provides a comprehensive understanding of the visual input, incorporating both local and global relationships.

\subsection{EmoPrompt Reasoning}
Chain-of-Thought (CoT)~\cite{cot} is a popular and efficient technique for enhancing the reasoning power of LLMs without fine-tuning. It involves adding step-by-step reasoning instructions to the user's prompt, guiding the LLM through a logical thought process. Given the delicate and unintuitive nature of emotional tasks, this kind of reasoning is crucial for accurate emotion understanding.

\begin{figure}[t]
\centering
\includegraphics[width=0.9\linewidth]{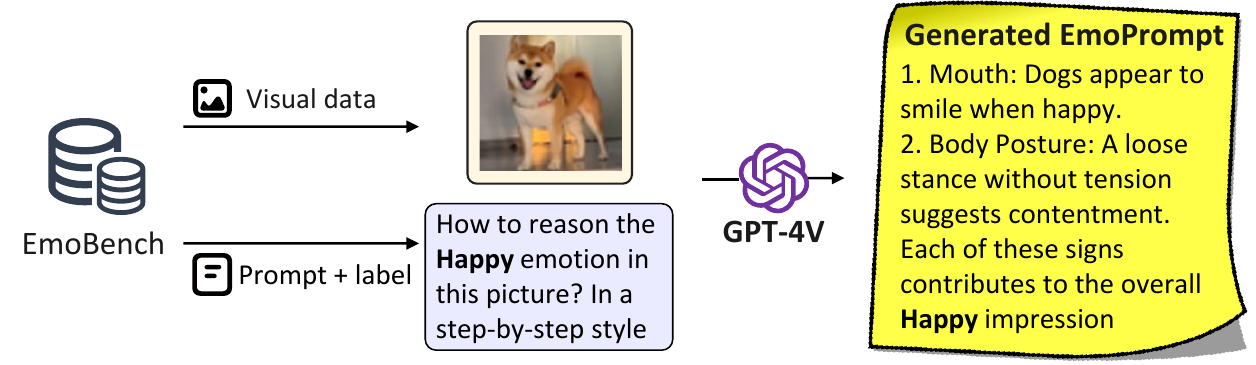}

\vspace{-0.8em}
\caption{Illustration of EmoPrompt. We utilize visual data and label pairs in EmoBench, and prompt GPT-4V~\cite{gpt4} to generate logical chains.}
\label{fig:fig3_emoprompt}\vspace{-1em}
\end{figure}

For emotional tasks, we first design a task-specific CoT as a baseline. Drawing inspiration from how humans identify emotions in images and videos, we observe that people often focus on the content of objects first, such as facial expressions, atmospheres, and other visual cues. Intuitively, we guide the MLLM to reason about the objective content in the data first, and then reason about the emotional task based on the obtained conclusion combined with the data. The advantage of this approach is that it guides the observation of LLM, leading to more robust reasoning.

However, this step-by-step thinking heavily depends on the observations made in the first step. If the LLM hallucinates or generates inaccurate observations during the initial stage, it can greatly affect the judgment of the emotional task. To address this issue, we propose EmoPrompt, which aims to provide correct guidance for the reasoning process.

To achieve this goal, we first collect data on a subset of emotional tasks along with their corresponding ground truth labels. By presenting both the ``question'' (emotion data) and ``answer'' (ground truth label) to GPT-4V, we obtain objective-to-subjective reasoning in the correct direction, as shown in ~\cref{fig:fig3_emoprompt}. This ensures the correctness of the step-by-step reasoning process. Using this methodology, we collect hundreds of examples of reasoning for each emotional task. These examples serve as demonstrations of correct reasoning during the EmoLLM reasoning process.

By incorporating EmoPrompt, we guide EmoLLM to follow a correct reasoning path, mitigating the impact of potential hallucinations or inaccuracies in the initial observation stage. This approach enhances the ability of LLMs to accurately understand and interpret emotions in multimodal data.

\begin{table}[t]
	\small	
	\centering	
	\caption
	{
        Comparison of the emotional ability between baseline MLLMs and our EmoLLM, on \textbf{EmoBench}-test set. }
 \begin{tabular}	{p{3.5cm}  |  >{\centering\arraybackslash}p{1cm} >{\centering\arraybackslash}p{1cm} >{\centering\arraybackslash}p{1cm}  >{\centering\arraybackslash}p{1cm}  >{\centering\arraybackslash}p{1cm} >{\centering\arraybackslash}p{1cm} >{\centering\arraybackslash}p{1cm} }
        \toprule
        \multirow{2}{*}{\makecell[l]{Methods}} & \multicolumn{7}{c}{EmoBench Testing (30K)} \\
        &Emo-C &  Emo-O &Intention  &Hate  &Humor  &Sarcasm & Overall \\
        \midrule
        Vicuna~\cite{vicuna} \hfill \textcolor{gray}{\textit{zero-shot}}& 29.21 & 21.55 & 17.48  & 45.39 & 49.68 & 55.23 & 28.63\\
        ChatUniVi~\cite{chatunivi}\hfill  \textcolor{gray}{\textit{fine-tune}}& 47.62 & 39.26 & 57.85 & 63.03 & 63.85 & 77.87 & 46.66 \\
        MacawLLM~\cite{lyu2023macaw} \hfill  \textcolor{gray}{\textit{fine-tune}}&42.42 & 31.05 & 52.91 & 57.54 & 55.60 & 71.75 & 40.28\\
        OneLLM~\cite{han2023onellm} \hfill  \textcolor{gray}{\textit{fine-tune}}&51.16 & 40.30 & 56.95 & 59.01 & 60.89 & 73.93 & 48.20\\
         \rowcolor{aliceblue!60}EmoLLM \hfill \textcolor{gray}{\textit{fine-tune}}& \textbf{64.06} & \textbf{52.58} & \textbf{73.99} & \textbf{67.43} & \textbf{75.69} & \textbf{86.67} & \textbf{60.36}\\
 	\bottomrule	
	\end{tabular}
 \vspace{-2em}
	\label{tbl:main_result}
\end{table}	

\vspace{-0.5em}
\section{Experiments}
\subsection{Experimental Setup}
We adopt CLIP (ViT-L/14)~\cite{clip} and WHISPER~\cite{whisper} as the visual and acoustic encoders, respectively. For the language foundation model, we choose the Vicuna-v1.5 model~\cite{vicuna}, which consists of 7B parameters. During the emotional fine-tuning stage, we utilize the data from \textbf{EmoBench}. EmoLLM is trained for 5 epochs with a batch size of 16, using the AdamW~\cite{adam1,adam2} optimizer with a cosine learning rate schedule. The learning rate is set to 2e-5, and the warmup rate is 0.03. All input images or frames are resized to 224 $\times$ 224. Training one epoch on 4 $\times$ RTX 4090 GPUs takes approximately 5 hours using LoRA~\cite{hu2021lora}. Hyperparameters are determined on the validation set, and final results are obtained on the test set. Each result is the average of three runs with various random seeds.

\begin{wraptable}{r}{6cm}
\centering
 \small
 \vspace{-2.0em}
 \caption
	{Comparison of the emotional ability between SOTA MLLMs and EmoLLM.}
 
 \vspace{-0.5em}
	\begin{tabular}{l|ccc}
    \toprule
    \label{tab:sotaemo}
     \textbf{Method}& \#Param &Emo-C & Emo-O \\\midrule
    
    \multirow{1}{*}{\shortstack{GPT-4V}} &$\sim{10}^{12}$ & 57.90 & 45.10 \\  
 \multirow{1}{*}{\shortstack{Gemini1.0}}&$\sim{10}^{11}$  & 45.47 & 44.83 \\
 \multirow{1}{*}{\shortstack{Gemini1.5}}&$\sim{10}^{11}$  & 45.47 & 44.83 \\
  \multirow{1}{*}{\shortstack{EmoLLM}}  &$\sim{10}^{10}$ &\textbf{75.03} &\textbf{67.14}\\\bottomrule
	\end{tabular}
 \vspace{-1em}
\end{wraptable}

\subsection{Main Results}
To quantitatively measure the emotional capability of EmoLLM, we evaluate its performance on six sub-tasks from EmoBench, including close-set and open-set emotion classification, intention recognition, and three special emotional application tasks. As shown in~\cref{tbl:main_result}, EmoLLM achieves superior performance compared to baselines with the same 7B parameter scale, demonstrating the effectiveness of our proposed approach.

We also compare the emotional understanding abilities of state-of-the-art MLLMs on an emotion sub-test set. Considering that some MLLMs do not support video and audio, we take a subset of pure images from EmoBench test set. It contains 6 emotion categories with hundreds of images in each category. As presented in~\cref{tab:sotaemo}, EmoLLM outperforms GPT-4V, Gemini-1.0, and Gemini-1.5 on both close-set (Emo-C) and open-set (Emo-O) emotion classification tasks while maintaining a smaller parameter count.

\subsection{Ablation Studies}
We conduct ablation studies to explore the key design choices in EmoLLM. All experiments are conducted on the Emo-C part of EmoBench test set, with other settings unchanged unless specified.

\begin{figure}
\centering
\includegraphics[width=0.9\linewidth]{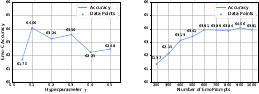}
\vspace{-0.5em}
 \caption{Hyperparameter ablation in Multi-perspective Visual Projection and EmoPrompts. EmoLLM has the best performance when $\tau$ is 0.1. For EmoPrompts, diversified prompts can enhance the emotional reasoning ability of LLM.}
\label{fig:fig4_abla}
\vspace{-1em}
\end{figure}

\noindent \textbf{Multi-perspective Visual Projection}
We investigate the impact of the hyperparameter $\tau$ in the Multi-perspective Visual Projection module by varying its value from 0.05 to 0.5. As shown in~\cref{fig:fig4_abla} (left), performance of EmoLLM is sensitive to the choice of $\tau$, with the highest accuracy of 64.06\% achieved when $\tau$ is set to 0.1. The accuracy tends to decline as $\tau$ increases, indicating that a suitable value of $\tau$ is beneficial for emotional understanding capabilities of EmoLLM.

\noindent \textbf{Quantity Effects in EmoPrompt}
To examine the impact of the number of EmoPrompts on the performance of EmoLLM, we vary the number of prompts from 100 to 1000 and evaluate the emotional capability. As depicted in~\cref{fig:fig4_abla} (right), increasing the number of EmoPrompts generally leads to improved performance, with the peak accuracy of 64.06\% achieved when all prompts are used. This finding highlights the importance of utilizing a diverse set of prompts to enhance the emotional reasoning ability of LLMs. However, the performance gains diminish as the number of prompts exceeds 600, suggesting an optimal range for balancing computational efficiency and emotional understanding.

\noindent \textbf{Effect of the Tuning Strategy}
We investigate whether different objective and affective training sequences affect the emotional understanding ability of LLMs. In~\cref{tab:training_strategy}, we compare the performance of three training strategies: \textbf{emo} (training with only EmoBench), \textbf{mix} (training with objective fine-tuned data mixed with EmoBench), and \textbf{sequential} (fine-tuning with objective data first and then with emotional task). The results suggest that sequential training substantially benefits emotional understanding. A possible explanation is that it simulates the way humans learn, starting with easy tasks and progressing to more difficult ones, while also moving from general knowledge to domain-specific knowledge.

\begin{table*}\small
    \centering
    \caption{Various training strategies affect emotional understanding ability of LLMs. Training on traditional tasks first and then emotional tasks (\textbf{\textit{sequential}}) leads to the best results.}
    
\vspace{-0.5em}
    \resizebox{\linewidth}{!}{\begin{tabular}{c|ccc|ccc|c}
\toprule
         Training Strategy  & \multirow{1}{*}{Emo-C}& \multirow{1}{*}{Emo-O} & \multirow{1}{*}{Intention} & \multirow{1}{*}{Hate}  & Humor   &Sarcasm   & \multirow{1}{*}{{Overall}} \\ \hline
     \textbf{\textit{emo}} & 61.65 & 47.40 & 67.71 & 62.44 & 70.82 & 80.22 & 56.24 \\

     \textbf{\textit{mix}} & 63.05$_\text{\textcolor{red}{+1.40}}$ & 49.32$_\text{\textcolor{red}{+1.92}}$ & 73.54$_\text{\textcolor{red}{+5.83}}$& 65.90$_\text{\textcolor{red}{+3.46}}$ & 67.65$_\text{\textcolor{gray}{-3.17}}$  & 83.74$_\text{\textcolor{red}{+3.52}}$ & 58.11$_\text{\textcolor{red}{+1.87}}$  \\
    \rowcolor{aliceblue!60}\textbf{\textit{sequential}} & 
64.06$_\text{\textcolor{red}{+2.41}}$ & 52.58$_\text{\textcolor{red}{+5.18}}$ & 73.99$_\text{\textcolor{red}{+6.28}}$ & 67.43$_\text{\textcolor{red}{+4.99}}$ & 75.69$_\text{\textcolor{red}{+4.87}}$  & 86.67$_\text{\textcolor{red}{+6.45}}$ &  60.36$_\text{\textcolor{red}{+4.12}}$  \\\bottomrule
    \end{tabular}}
    \vspace{-6pt}
    \label{tab:training_strategy}
\end{table*}

\vspace{-0.5em}
\section{Conclusion}
In this work, we introduce EmoBench, a comprehensive benchmark for enhancing and evaluating the emotional understanding capabilities of Multimodal Large Language Models (MLLMs), and propose EmoLLM, a novel model incorporating Multi-perspective Visual Projection and EmoPrompt techniques. Through extensive experiments on EmoBench, we demonstrated substantial improvements of EmoLLM over baselines, with an average improvement of 12.1\% across multiple foundation models.

\noindent \textbf{Limitations.}
One notable limitation is that the answers to the instructions in EmoBench may lack diversity since they were generated by GPT-4 and automated scripts rather than collected from human annotators. Maybe the future of work combining automation with manual labeling is a promising direction. Another limitation is the inherent vulnerabilities of LLMs, such as hallucination and sensitivity to prompts, which may affect the performance of EmoLLM. 

\noindent \textbf{Future Work.}
Despite these limitations, we believe our work takes a significant step towards enabling MLLMs to achieve a deeper understanding of complex emotions in multimodal data, paving the way for emotionally intelligent AI systems. Future work could focus on addressing the limitations mentioned above, such as increasing the diversity of EmoBench through a combination of automated and manual labeling, and mitigating the vulnerabilities of LLMs. Furthermore, exploring the application of emotionally intelligent AI systems in real-world scenarios and evaluating their impact on user experience and well-being could be valuable avenues for future research.


\bibliographystyle{IEEEtran}
\bibliography{ref}

\medskip

\end{document}